\documentclass[letterpaper, 10 pt, conference]{IEEEtran}  
\IEEEoverridecommandlockouts                              
\usepackage[english]{babel}

\usepackage[T1]{fontenc} 
\usepackage[cmex10]{amsmath}
\usepackage{amsfonts}
\usepackage{pifont}
\usepackage{cite} 
\usepackage{fixltx2e}
\usepackage[pdftex]{graphicx, color}
\usepackage{caption}
\usepackage{subcaption}
\usepackage{paralist}
\usepackage{hyperref}
\graphicspath{{../media/diagrams/}{../media/}{./}}

\usepackage{color,soul}  

\newcommand{\mediaprefix}[1]{}
\newcommand{\pdftexprefix}{./}
\newcommand{\pdftexprefixdiagrams}{./}
\newcommand{\tickYes}{\checkmark}
\newcommand{\tickNo}{\hspace{1pt}\ding{55}}

\newcommand{\vect}[1]{\mathbf{#1}}
\newcommand{\hvect}[1]{\bar{\vect{#1}}}
\newcommand{\uvect}[1]{\hat{\vect{#1}}}

\newcommand{\improvementfolds}{ten}
\newcommand{\meanartktranserr}{0.57m}
\newcommand{\meanmutloctranserr}{0.016m}
\newcommand{\meanbundlertranserr}{0.20m}
\newcommand{\meanartkroterr}{$9.2^\circ$}
\newcommand{\meanmutlocroterr}{$0.33^\circ$}
\newcommand{\meanbundlerroterr}{$0.016^\circ$}
\DeclareMathOperator{\Tr}{Tr}

\title{\LARGE \bf
Mutual Localization: Two Camera Relative 6-DOF Pose Estimation from Reciprocal 
Fiducial Observation
}

\author{Vikas Dhiman, Julian Ryde, Jason J. Corso
\thanks{J.J. Corso, J. Ryde and Vikas Dhiman are with Department of Computer Science and Engineering,
        SUNY at Buffalo, Buffalo, NY, USA 
        {\tt\small \{jcorso,jryde,vikasdhi\}@buffalo.edu}}%
    }

\begin{document}

\maketitle
\thispagestyle{empty}
\pagestyle{empty}

\begin{abstract}

Concurrently estimating the 6-DOF pose of multiple cameras or robots---cooperative localization---is a core problem in contemporary robotics.  Current works focus on a set of mutually observable world landmarks and often require inbuilt egomotion estimates; situations in which both assumptions are violated often arise, for example, robots with erroneous low quality odometry and IMU exploring an unknown environment.  In contrast to these existing works in cooperative localization, we propose a cooperative localization method, which we call \textit{mutual localization}, that uses reciprocal observations of camera-fiducials to obviate the need for egomotion estimates and mutually observable world landmarks.  We formulate and solve an algebraic formulation for the pose of the two camera mutual localization setup under these assumptions.  Our experiments demonstrate the capabilities of our proposal egomotion-free cooperative localization method: for example, the method achieves 2cm range and 0.7 degree accuracy at 2m sensing for 6-DOF pose.  To demonstrate the applicability of the proposed work, we deploy our method on Turtlebots and we compare our results with ARToolKit\cite{kato1999} and Bundler\cite{snavely2006photo}, over which our method achieves a \improvementfolds fold improvement in translation estimation accuracy.

\end{abstract}

\section{Introduction}






Cooperative localization is the problem of finding the relative 6-DOF pose 
between robots using sensors from more than one robot. Various strategies 
involving different sensors have been used to solve this problem.  For example, 
Cognetti et al. \cite{cognetti2012mutual, franchi2009mutual} use multiple
bearning-only observations with a motion detector to solve for cooperative
localization among multiple anonymous robots. Trawny et al.  \cite{trawny2010interrobot}
and lately Zhou et al. \cite{zhou2010determining, zhou2012determining} 
provide a comprehensive mathematical analysis of solving cooperative 
localization for different cases of sensor data availability.
Section \ref{sec:related} covers related literature in more detail.

To the best of our knowledge, all other cooperative localization works (see
Section \ref{sec:related}) require estimation of
egomotion.  However, a dependency on egomotion is a limitation for systems
that do not have gyroscopes or accelerometers, which can provide displacement
between two successive observations. Visual egomotion, like MonoSLAM
\cite{davison2007monoslam}, using distinctive image
features estimates requires high quality correspondences, which remains a
challenge in machine vision, especially in cases of non-textured environments.
Moreover, visual egomotion techniques are only correct upto a scale factor.
Contemporary cooperative localization methods that use egomotion
\cite{zhou2010determining, trawny2010interrobot, martinelli2012vision} yield
best results only with motion perpendicular to the direction of mutual
observation and fails to produce results when either observer
undergoes pure rotation or motion in the direction of observation.
Consequently, in simple robots like Turtlebot,
this technique produces poor results because of absence of sideways motion
that require omni-directional wheels. 

\begin{figure}
  \def\svgwidth{0.5\textwidth}
  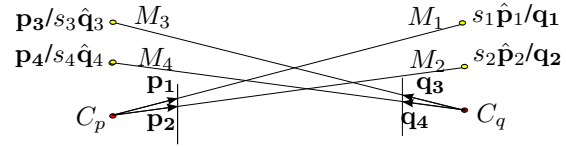
  \caption{Simplified diagram for the two-camera problem. Assuming the length
  of respective rays to be $s_1, s_2, s_3, s_4$ respectively, each marker
coordinates can be written in both coordinate frames $\{p\}$ and $\{q\}$. For
example $M_1$ is $s_1\uvect{p}_1$ in frame $\{p\}$ and $\vect{q_1}$ in
$\{q\}$, where $\uvect{p}_1$ unit vector parallel to $\vect{p}_1$.}
  \label{fig:2camprob}
\end{figure}

To obviate the need for egomotion, we propose a method for relative pose  
estimation that leverages distance between fiducial markers mounted on
robots for resolving scale ambiguity.  Our method, which we call 
\textit{mutual localization}, depends upon the simultaneous mutual/reciprocal 
observation of bearing-only sensors.  Each sensor is outfitted 
with fiducial markers (Fig. \ref{fig:2camprob}) whose position within the host
sensor coordinate system is known, in contrast to assumptions in earlier works that
multiple world landmarks would be concurrently observable by each sensor
\cite{zou2012coslam}. Since our method does not depend on egomotion, hence it
is instantaneous, which means it is robust to false negatives and it do not
suffers from the errors in egomotion estimation. 

The main contribution of our work is a generalization of
\emph{Perspective-3-Points} (P3P) problem where observer and the observed
points are distributed in different reference frames unlike conventional approach
where observer's reference frame do not contain any observed points and vice
versa.
In this paper we present an algebraic derivation to solve for the
relative camera pose (rotation and translation) of the two bearing-only
sensors in the case that each can observe two known fiducial points in the
other sensor; essentially giving an algebraic system to compute the relative
pose from four correspondences (only three are required in our algorithm but
we show how the fourth correspondence can be used to generate a set of hypothesis
solutions from which best solution can be chosen).
Two fiducial points on each robot (providing four correspondences) are preferable
to one on one and two on the other, as it allows extension to multi-robot
($>2$) systems ensuring that any pair of similarly equipped robots can
estimate their relative pose. In this paper, we focus on only two robot case
as an extension to multi-robot case as pairwise localization is trivial yet
practically effective.

Our derivation, although inspired by the linear pose estimation method of Quan
and Lan \cite{quan1999linear}, is novel since all relevant past works we know
on P3P problem \cite{haralick1994review},
assume all observations are made in one coordinate frame and observed points
in the other.  In contrast, our
method makes no such assumption and concurrently solves the pose estimation
problem for landmarks sensed in camera-specific coordinates frames.

We demonstrate the effectiveness of our method, by analyzing its
accuracy in both synthetic, which affords quantitative absolute assessment,
and real localization situations by deployment on Turtlebots. 
%
We use 3D reconstruction experiments to show the accuracy of our algorithm. 
Our experiments demonstrate the effectiveness of the proposed approach.



\section{Related Work}
\label{sec:related}

Cooperative localization has been extensively studied and applied to
various applications. One of the latest works in this area comes from Cognetti et al.
\cite{cognetti2012mutual}, \cite{franchi2009mutual} where they focus on the
problem of cooperatively localizing multiple robots anonymously. They use
multiple bearing-only observations and a motion detector to localize the
robots. The robot detector is a simple feature extractor that detects vertical
cardboard squares mounted atop each robot in the shadow zone of the range
finder.  One of oldest works come from Karazume
et. al.  \cite{kurazume1994cooperative} where they focus on using cooperative
localization as a substitute to dead reckoning by suggesting a ``dance'' in
which robots act as mobile landmarks. Although they do not use egomotion, but
instead assume that position of two robots are known while localizing the
third robot. Table \ref{tab:relatedwork} summarizes a few closely related works
with emphasis on how our work is different different from each of them. Rest
of the section discusses those in detail.

Howard et al. \cite{howard1999cooperative} coined the CLAM (Cooperative
Localization and Mapping) where they concluded that as an observer robot observes the
explorer robot, it improves the localization of robots by the new constraints of
observer to explorer distance. Recognizing that odometry errors can cumulate
over time, they suggest using constraints based on cooperative localization to
refine the position estimates. Their approach, however, do not utilizes the
merits of mutual observation as they propose that one robot explores the
world and other robot watches. We show in our experiments, by comparison to
ARToolKit \cite{kato1999} and Bundler \cite{snavely2006photo}, that mutual
observations of robots can be up to 10 times more accurate than observations
by single robot.

A number of groups have considered cooperative vision and laser based mapping in outdoor environments \cite{madhavan2004, ryde2006} and vision only \cite{little1998, rocha2005}. Localization and mapping using heterogeneous robot teams with sonar sensors is examined extensively by \cite{grabowski2001, grabowski2003}. Using more than one robot enables easier identification of previously mapped locations, simplifying the loop-closing problem \cite{konolige1999}.
%
%
%

Fox et al. \cite{fox1999collaborative} propose cooperative localization based
on Monte-Carlo localization technique.  The method uses odometry measurements
for ego motion.  Chang et al. \cite{chang2011vision} uses depth and
visual sensors to localize Nao robots in the 2D ground plane.
Roumeliotis and Bekey \cite{roumeliotis2002distributed} focus on sharing
sensor data across robots, employing as many sensors as
possible which include odometry and range sensors.
Rekleitis et al. \cite{rekleitis97multi-robotexploration} provide a model of
robots moving in 2D equipped with both distance and bearing sensors.

\begin{table}
  \begin{tabular}{|l|c|c|c|c|c|}
    \hline
    Related work $\backslash$ Tags                                & NoEM     & BO       & NoSLAM  & MO\\
    \hline
    Mutual localization                                           & \tickYes & \tickYes & \tickYes & \tickYes\\
    Howard et al.\cite{howard1999cooperative}                     & \tickNo  & \tickYes & \tickYes & \tickYes\\
    Zou and Tan \cite{zou2012coslam}                              & \tickYes & \tickYes & \tickNo  & \tickNo \\
    Cognetti et al.\cite{cognetti2012mutual}                      & \tickNo  & \tickYes & \tickYes & \tickNo \\
    Trawny et al.\cite{trawny2010interrobot}                      & \tickNo  & \tickYes & \tickYes & \tickYes\\
    Zhou and Roumeliotis \cite{zhou2010determining, zhou2012determining}               & \tickNo  & \tickYes & \tickYes & \tickYes\\
    Roumeliotis et al.\cite{roumeliotis2002distributed}           & \tickNo  & \tickNo  & \tickNo  & \tickYes\\
    %
    %
    \hline
  \end{tabular}
  \\
  \\
  where\\
  \\
  \begin{tabular}{|l|p{0.7\columnwidth}|}
    \hline
    Tag & meaning\\
    \hline
    NoEM & Without Ego-Motion. All those works that use egomotion are marked
    as \tickNo. \\
    BO & Localization using bearing only measurements. No depth measurements
    required. All those works that require depth measurements are marked with
    \tickNo. \\
    NoSLAM & SLAM like tight coupling. Inaccuracy
    in mapping leads to cumulating interdependent errors in localization and
    mapping. All those works that use SLAM like approach are marked with a
    \tickNo\\
    MO & Utilizes mutual observation, which is more accurate than one-sided
    observations. All those works that do not use mutual observation, and
    depend on one-sided observations are marked as \tickNo \\
    \hline
  \end{tabular}
  \caption{Comparison of related work with Mutual localization}
  \label{tab:relatedwork}
\end{table}

Zou and Tan \cite{zou2012coslam} proposed a cooperative
simultaneous localization and mapping method, CoSLAM, in which multiple robots 
concurrently observe the same scene.  Correspondences in time (for each robot) 
and across robots are fed into an extended Kalman filter and used to 
simultaneously solve the localization and mapping problem.  However, this and 
other ``co-slam'' approaches such as \cite{kim2005vision} remain limited due to 
the interdependence of localization and mapping variables: errors in the map are 
propagated to localization and vice versa.

Recently Zhou and Roumeliotis \cite{zhou2010determining,
zhou2012determining} have published solution of a set of 14 minimal solutions
that covers a wide range of robot to robot measurements. However, they use
egomotion for their derivation and they assume that observable
fiducial markers coincide with the optical center of the camera. Our work does
not make any of the two assumptions.

\section{Problem Formulation}

We use the following notation in this paper, see Fig. \ref{fig:2camprob}.
    $C_p$ and $C_q$ represent two robots, each with a camera as a sensor.
    The corresponding coordinate frames are $\{p\}$ and $\{q\}$
    respectively with origin at the optical center of the camera.
    Fiducial markers $M_1$ and $M_2$ are fixed on robot $C_q$ and hence their
    positions are known in frame $\{q\}$ as $\vect{q}_1, \vect{q}_2 \in \mathbb{R}^3$.
    Similarly, $\vect{p}_3, \vect{p}_4 \in \mathbb{R}^3$ are the positions of
    markers $M_3$ and $M_4$ in coordinate frame $\{p\}$.
    Robots are positioned such that they can observe each others markers in
    their respective camera sensors.
    The 2D image coordinates of the markers $M_1$ and $M_2$ in the image
    captured by the camera $\{p\}$ are measured as $\hvect{p}_1, \hvect{p}_2 \in
    \mathbb{R}^2$ and that of $M_3$ and $M_4$ is $\hvect{q}_3,
    \hvect{q}_4 \in \mathbb{R}^2$ in camera $\{q\}$.
    Let $K_p, K_q \in \mathbb{R}^{3\times3}$ be the intrinsic camera
    matrices of the respective camera sensors on robot $C_p, C_q$.
    Also, note the superscript notation. 2D image coordinates are
    denoted by a \emph{bar}, example $\hvect{p}$. Unit vectors that provide
    bearing information are denoted by a \emph{caret}, example $\uvect{p}$.

Since the real life images are noisy, the measured image positions $\hvect{p}_i$
and $\hvect{q}_i$ will differ from the actual positions $\hvect{p}_{i0}$ and
$\hvect{q}_{i0}$ by gaussian noise $\vect{\eta}_{i}$.
\begin{eqnarray}
  \hvect{p}_i &= \hvect{p}_{i0} + \vect{\eta}_{pi} & \forall i \in \{1, 2\} \\
  \hvect{q}_i &= \hvect{q}_{i0} + \vect{\eta}_{qi} & \forall i \in \{3, 4\}
\end{eqnarray}
The problem is to determine the rotation $R \in \mathbb{R}^{3\times3}$
and translation $\vect{t} \in \mathbb{R}^3$ from frame $\{p\}$ to frame
$\{q\}$ such that any point $\vect{p}_i$ in frame $\{p\}$ is related to its
corresponding point $\vect{q}_i$ in frame $\{q\}$ by the following equation.
\begin{align}
  \vect{q}_i =R \vect{p}_i + \vect{t}
\end{align}
The actual projections of markers $M_i$ into the camera image frames of the
other robot are governed by following equations,
\begin{eqnarray}
\hvect{p}_{i0} &= f(K_pR^{-1}(\vect{q}_i  - \vect{t})) & \forall i \in \{1, 2\}\\
    \hvect{q}_{i0} &= f(K_q(R\vect{p}_i + \vect{t}))  & \forall i \in \{3, 4\}
\end{eqnarray}
where $f$ is the projection function defined over a vector \\$\vect{v} =
\begin{bmatrix} v_x, v_y, v_z \end{bmatrix}^\top$ as 
\begin{align}
  f(\vect{v}) = \begin{bmatrix} \frac{v_x}{v_z}, \frac{v_y}{v_z} \end{bmatrix}^\top
\end{align}
To minimize the effect of noise we must compute the optimal transformation,
$R^*$ and $\vect{t}^*$.
\begin{multline}
    (R^*, \vect{t}^*) = \arg \min_{(R,t)}\left(
    \sum_{i \in \{1, 2\}} \| \hvect{p}_i - f(K_pR^{-1}(\vect{q}_i - \vect{t})) \|^2\right. \\
                    \left.+ \sum_{i \in \{3, 4\}} \| \hvect{q}_i -
    f(K_q(R\vect{p}_i + \vect{t})) \|^2\right)
  \label{eq:proboptimization}
\end{multline}

To solve this system of equations we start with exact equations that lead to a
large number of polynomial roots. To choose the best root among the set of
roots we use the above minimization criteria.

Let $\uvect{p}_i, \uvect{q}_i \in \mathbb{R}^3$ be the unit vectors
drawn from the camera's optical center to the image projection of the
markers. 
The unit vectors can be computed from the position
of markers in camera images $\hvect{p}_i, \hvect{q}_i$ by the following
equations.
\begin{eqnarray}
  \uvect{p}_i &= \frac{K_p^{-1}\begin{bmatrix}\hvect{p}_i^\top, 
1\end{bmatrix}^\top}{\|K_p^{-1}\begin{bmatrix}\hvect{p}_i^\top, 1\end{bmatrix}^\top\|} & \forall i \in \{1, 2\}\\
  \uvect{q}_i &= \frac{K_q^{-1}\begin{bmatrix}\hvect{q}_i^\top, 
1\end{bmatrix}^\top}{\|K_q^{-1}\begin{bmatrix}\hvect{q}_i^\top,
1\end{bmatrix}^\top\|} & \forall i \in \{3, 4\}
\end{eqnarray}

Further let $s_1$, $s_2$ be the distances of markers $M_1$, $M_2$ from the optical
center of the camera sensor in robot $C_p$. And $s_3$, $s_4$ be the distances of
markers $M_3$, $M_4$ from the optical center of camera sensor in robot $C_q$.
Then the points $\vect{q}_1$, $\vect{q}_2$, $s_3\uvect{q}_3$, $s_4\uvect{q}_4$ in
coordinate frame $\{q\}$ correspond to the points $s_1\uvect{p}_1$,
$s_2\uvect{p}_2$, $\vect{p}_3$, $\vect{p}_4$ in coordinate frame $\{p\}$.
\begin{align}
  \begin{split}
      \vect{q}_1     &= t + s_1R \uvect{p}_1\\
      \vect{q}_2     &= t + s_2R \uvect{p}_2\\
      s_3\uvect{q}_3 &= t+ R \vect{p}_3\\
      s_4\uvect{q}_4 &= t + R \vect{p}_4
  \end{split}
  \label{eq:4RTeq}
\end{align}
These four vector equations provide us 12 constraints (three for each
coordinate in 3D) for our 10 unknowns (3 for rotation $R$,
3 for translation $t$, and 4 for $s_i$). We first consider only the
first three equations, which allows an exact algebraic solution of the nine
unknowns from the nine constraints.  

Our approach to solving the system is inspired by the well studied
problem of \emph{Perspective-3-points} \cite{haralick1994review}, also
known as \emph{space resection} \cite{quan1999linear}. However, note that the
method cannot be directly applied to our problem as known points are distributed in
both coordinate frames as opposed to the space resection problem where all
the known points are in the one coordinate frame.

The basic flow steps of our approach are to first solve for the
three range factors, $s_1, s_2$ and $s_3$ (Section \ref{sec:solvescale}). 
Then we set up a classical absolute orientation system on the rotation and
translation (Section \ref{sec:classical}), which is solved using established
methods such as Arun et al. \cite{arun1987least} or Horn \cite{horn1987closed};
finally, since our algebraic solution will give rise to many candidate roots,
we develop a root-filtering approach to determine the best solution (Section
\ref{sec:filtering}). 

\subsection{Solving for $s_1$, $s_2$ and $s_3$}
\label{sec:solvescale}

The first step is to solve the system for $s_1, s_2$ and $s_3$.  We eliminate
$R$ and $t$ by considering the inter-point distances in both coordinate frames.
\begin{align}
  \begin{split}
    \label{eq:solvescale}
    \|s_1\uvect{p}_1-s_2\uvect{p}_2\| &=\|   \vect{q_1}-   \vect{q_2}\| \\
    \|s_2\uvect{p}_2-   \vect{p_3}\|  &=\|   \vect{q_2}-s_3\uvect{q}_3\| \\
    \|   \vect{p_3}-s_1\uvect{p}_1\|  &=\|s_3\uvect{q}_3-   \vect{q_1}\|
  \end{split}
\end{align}
Squaring both sides and representing the vector norm as the dot product 
gives the following system of polynomial equations.
\begin{subequations}
  \label{eq123}
  \begin{align}
    s_1^2 + s_2^2 	  & - 2s_1s_2\uvect{p}_1^\top\uvect{p}_2 - \|\vect{q}_1 - 
    \vect{q}_2\|^2 = 0 \label{eq12}\\
    s_2^2 - s_3^2 & - 2s_2\uvect{p}_2^\top\vect{p}_3
              + 2s_3\vect{q}_2^\top\uvect{q}_3
          + \|\vect{p}_3\|^2 - \|\vect{q}_2\|^2 = 0
          \label{eq23}\\
    s_1^2          - s_3^2 & - 2s_1\uvect{p}_1^\top\vect{p}_3
              + 2s_3\vect{q}_1^\top\uvect{q}_3
          + \|\vect{p}_3\|^2 - \|\vect{q}_1\|^2 = 0
          \label{eq31}
  \end{align}
\end{subequations}
This system has three quadratic equations implying a Bezout bound of eight
($2^3)$ solutions. 
Using the Sylvester resultant we sequentially eliminate variables from each
equation. Rewriting \eqref{eq12} and \eqref{eq23} as quadratics in terms of
$s_2$ gives
\begin{align}
  \label{eq:quad_s2}
  s_2^2 + \underbrace{(-2s_1\uvect{p}_1^\top\uvect{p}_2)}_{a_1}s_2 +
  \underbrace{(s_1^2 - |q_1 - q_2|^2)}_{a_0} = 0 \\
  \label{eq:quad_s22}
  s_2^2 + \underbrace{(-2\uvect{p}_2^\top\vect{p}_3)}_{b_1}s_2 - \underbrace{(s_3^2 -
  2s_3\vect{q}_2^\top\uvect{q}_3 - \|\vect{p}_3\|^2 + \|\vect{q}_2\|^2)}_{b_0} = 0
\end{align}
The Sylvester determinant \cite[p.~123]{bykov1998elimination} of
\eqref{eq:quad_s2} and \eqref{eq:quad_s22} is given by the determinant of the
matrix formed by the coefficients of $s_2$.
\begin{align}
  r(s_1, s_3) = 
  \begin{vmatrix}
    1 & a_1 & a_0 & 0 \\
    0 & 1 & a_1 & a_0 \\
    1 & b_1 & b_0 & 0 \\
    0 & 1 & b_1 & b_0
  \end{vmatrix}
  \label{eq:sylvester}
\end{align}
This determinant is a quartic function in $s_1$,
$s_3$. By definition of resultant, the resultant is zero if and only if the parent
equations have at least a common root\cite{bykov1998elimination}. Thus we have
eliminated variable $s_2$ from \eqref{eq12} and \eqref{eq23}. We can repeat
the process for eliminating $s_3$ by rewriting $r(s_1, s_3)$ and \eqref{eq31} as:
\begin{equation}
  \begin{split}
    r(s_1, s_3) = c_4s_3^4 + c_3s_3^3 + c_2s_3^2 + c_1s_3 + c_0 &= 0\\
        -s_3^2 + \underbrace{(2\vect{q}_1^\top\uvect{q}_3)}_{d_1}s_3
        + \underbrace{s_1^2 - 2s_1\uvect{p}_1^\top\vect{p}_3
        + \|\vect{p}_3\|^2 - \|\vect{q}_1\|^2}_{d_0} &= 0
    \label{eq:s2eliminated}
  \end{split}
\end{equation}
The Sylvester determinant of \eqref{eq:s2eliminated} would be
\begin{align}
  r_2(s_1) = \begin{vmatrix}
    c_4 & c_3 & c_2 & c_1 & c_0 & 0 \\
    0  & c_4 & c_3 & c_2 & c_1 & c_0\\
   1 & d_1 & d_0 & 0 & 0 & 0 \\
     0 & 1 & d_1 & d_0 & 0 & 0 \\
     0 & 0 & 1 & d_1 & d_0 & 0 \\
     0 & 0 & 0 & 1 & d_1 & d_0
  \end{vmatrix} = 0.
  \label{eq:finalsylvester}
\end{align}
Solving \eqref{eq:finalsylvester} gives an 8 degree polynomial in $s_1$.
By Abel-Ruffini theorem \cite[p.~131]{barbeau2003polynomials}, a closed-form
solution of the above polynomial does not exist.

The numeric solution to \eqref{eq:finalsylvester} gives eight roots for $s_3$. We 
compute $s_1$ and $s_2$ using \eqref{eq31} and \eqref{eq23} respectively. 
Because the camera cannot see objects behind it, only real
positive roots are maintained from the resultant solution set. 

\subsection{Solving for $R$ and $t$}
\label{sec:classical}

With the solutions for the scale factors, $\{s_1, s_2, s_3\}$ we can
compute the absolute location of the Markers $\{M_1, M_2, M_3\}$ in both the
frames $\{p\}$ and $\{q\}$.
\begin{eqnarray*}
  \vect{p}_i = s_i\uvect{p}_i &\forall i \in \{1, 2\}\\
  \vect{q}_i = s_i\uvect{q}_i &\forall i \in \{3\}
\end{eqnarray*}
These exact correspondences give rise to the
classical problem 
of absolute orientation i.e. given three points in two coordinate frames find
the relative rotation and translation between the frames.
For each positive root of $s_1$, $s_2$, $s_3$ we use the method in 
Arun et. al \cite{arun1987least} method (similar to Horn's method
\cite{horn1987closed}) to compute the corresponding rotation $R$ and translation
value $\vect{t}$.

\subsection{Choosing the optimal root}
\label{sec:filtering}

Completing squares in \eqref{eq123} yields important information about
redundant roots.
\begin{subequations}
  \label{eq:squares}
  \begin{align}
  (s_1 + s_2)^2& - 2s_1s_2(1 + \uvect{p}_1^\top\uvect{p}_2) - \|\vect{q}_1 - 
  \vect{q}_2\|^2 = 0
  \end{align}
  \begin{multline}
  (s_2 - \uvect{p}_2^\top\vect{p}_3)^2
    - (s_3 - \vect{q}_2^\top\uvect{q}_3)^2\\
    + (\vect{p}_3 - \uvect{p}_2)^\top\vect{p}_3 
    - \vect{q}_2^\top(\vect{q}_2 - \uvect{q}_3) = 0\\
  \end{multline}
  \begin{multline}
  (s_1 - \uvect{p}_1^\top\vect{p}_3)^2
    - (s_3 - \vect{q}_1^\top\uvect{q}_3)^2\\
    + (\vect{p}_3 - \uvect{p}_1)^\top\vect{p}_3 
    - \vect{q}_1^\top(\vect{q}_1 - \uvect{q}_3) = 0\\
  \end{multline}
\end{subequations}

Equations \eqref{eq:squares} do not put any constraints on positivity of terms
$(s_2 - \uvect{p}_2^\top\vect{p}_3)$, $(s_3 - \vect{q}_2^\top\uvect{q}_3)$,
$(s_1 - \uvect{p}_1^\top\vect{p}_3)$ or $(s_3 - \vect{q}_1^\top\uvect{q}_3)$.
However, all these terms are positive as long as the markers of the observed robot 
are farther from the camera than the markers of the
observing robot.  Also, the distances $s_i$ are
assumed to be positive. Assuming the above, we filter the
\emph{real} roots by the following criteria:
\begin{align}
  s_1 &\ge \|\vect{p}_3\|\\
  s_2 &\ge \|\vect{p}_3\|\\
  s_3 &\ge \max(\|\vect{q}_1\|, \|\vect{q}_2\|)
\end{align}
These criteria not only reduce the number of roots significantly, but also
filter out certain degenerate cases.

For all the filtered roots of \eqref{eq:finalsylvester}, we compute
the corresponding values of $R$ and $\vect{t}$, choosing the best
root that minimizes the error function, \eqref{eq:proboptimization}.

\subsection{Extension to more than three markers}
%

Even though the system is solvable by only three markers, we choose to use
four 
markers for symmetry. We can fall back to the three marker solution in situations 
when one of the markers is occluded.
Once we extend this system to 4 marker points, we obtain 6 bivariate quadratic
equations instead of the three in \eqref{eq123} that can be reduced to three 8-degree
univariate polynomials.
The approach to finding the root with the least error is the same as described above. 

The problem of finding relative pose from five or more markers is better
addressed by solving for the homography when two cameras observe
the same set of points as done by  \cite{stewenius2006recent, nister2004efficient,
philip1996non-iterative, longuet1987computer}. The difference for us is that the 
distance between the points in both coordinate frames is known hence we can
estimate the translation metrically which is not the case in classical
homography estimation. Assuming the setup for five points such that 
\eqref{eq:4RTeq} becomes
\begin{align}
  \begin{split}
      \vect{q}_1     &= t + s_1R \uvect{p}_1\\
      \vect{q}_2     &= t + s_2R \uvect{p}_2\\
      s_3\uvect{q}_3 &= t + R \vect{p}_3\\
      s_4\uvect{q}_4 &= t + R \vect{p}_4\\
      s_5\uvect{q}_5 &= t + R \vect{p}_5
  \end{split}
  \label{eq:5RTeq}
\end{align}

If the essential matrix is $E$, the setup is the same as solving for 
\begin{align}
  [\vect{q}_1 , \vect{q}_2 , \uvect{q}_3 , \uvect{q}_4 , \uvect{q}_5 ]^\top
  E
  [\uvect{p}_1 , \uvect{p}_2 , \vect{p}_3 , \vect{p}_4 , \vect{p}_5 ] = 0
  \label{eq:essentialmatrix}
\end{align}
The scale ambiguity of the problem can be resolved by one of the distance relations
from \eqref{eq:solvescale}. Please refer to \cite{nister2004efficient} for
solving \eqref{eq:essentialmatrix}. For more points refer to \cite{hartley2000multiple} for the 
widely known 7-point and linear 8-point algorithms.



\section{Implementation}

\begin{figure}
  \includegraphics[width=\columnwidth]{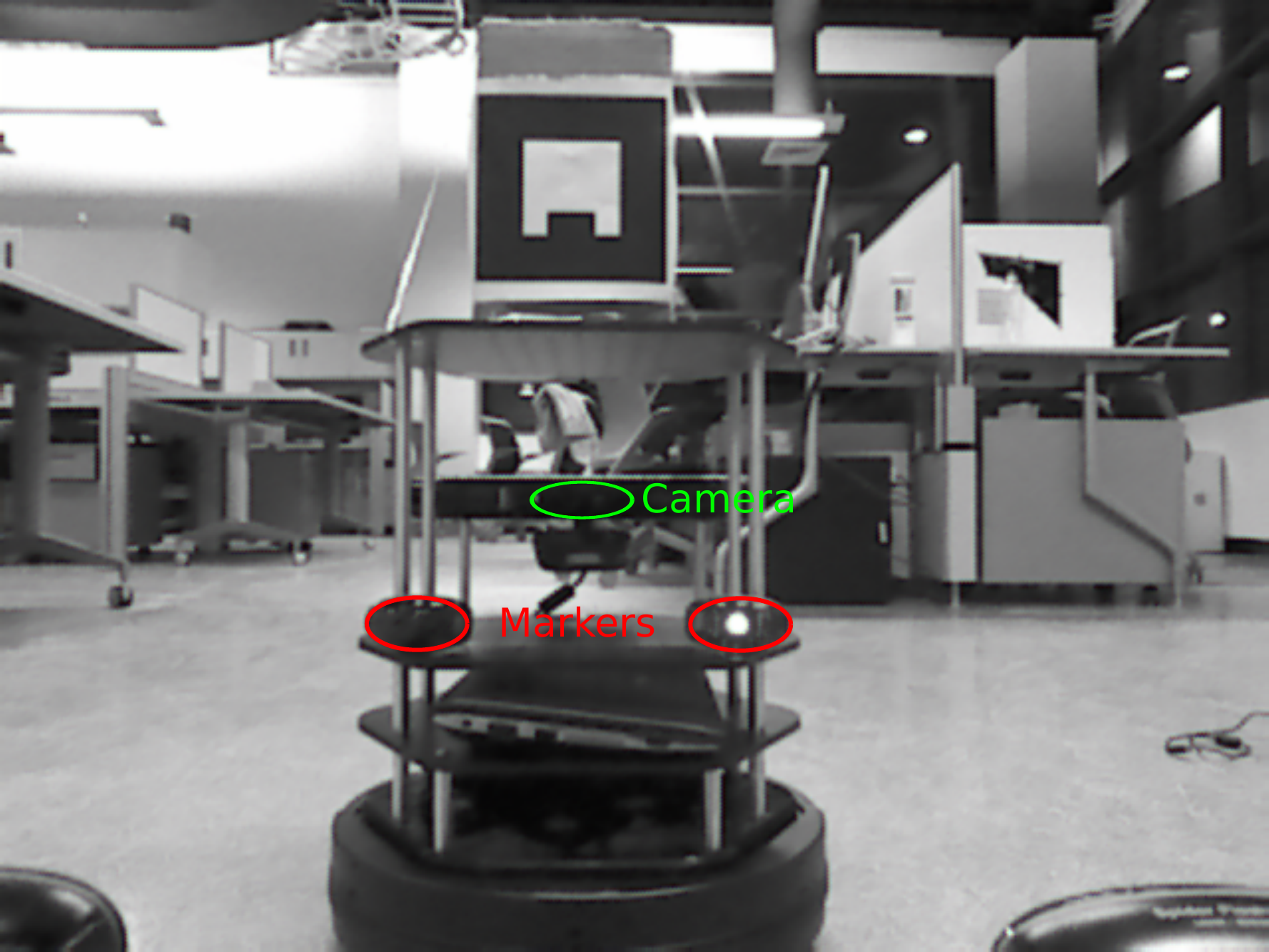}
  \caption{The deployment of markers on Turtlebot that we used for our
  experiments}
  \label{fig:turtlebotdeployment}
\end{figure}

We implement our algorithm on two Turtlebots with fiducial markers.
One of the Turtlebots with markers is shown in Fig. \ref{fig:turtlebotdeployment}.
We have implemented the algorithm in Python using the Sympy
\cite{certik2008sympy}, OpenCV \cite{bradski2000opencv}
and Numpy \cite{developers2010scientific} libraries. As the implementing
software formulates and solves polynomials symbolically, it is generic enough
to handle any reasonable number of points in two camera coordinate frames. We
have tested the solver for the following combination of points: 0-3, 1-2, 2-2,
where 1-2 means that 1 point is known in the first coordinate frame and 2 points are known in the second.

We use blinking lights as fiducial markers on the robots and barcode-like
cylindrical markers as for the 3D reconstruction experiment.

\begin{figure*}
    \def\svgwidth{\textwidth}
    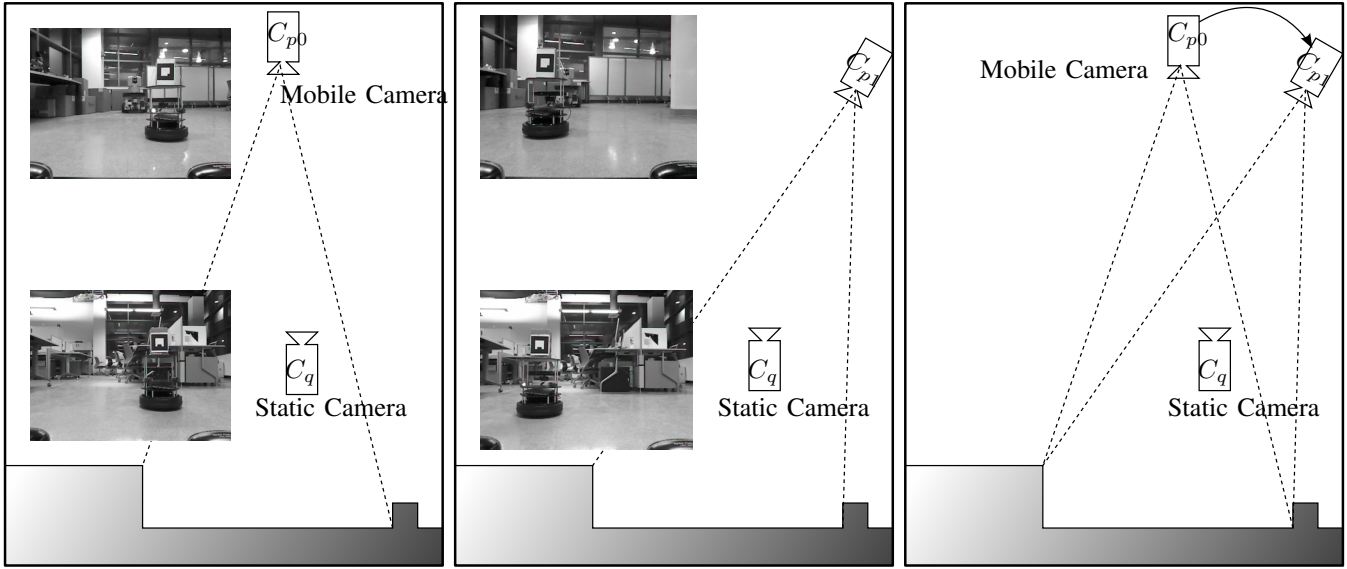
    \caption{Diagram of the two camera setup for mutual localization 3D metric
    reconstruction, along with images from each camera for two poses of the mobile
    camera.  Cameras have distinctive cylindrical barcode-like markers to aid
    detection in each others image frames.  Also depicted is the triangulation to
    two example feature points.}
    \label{fig:setupdiagram}
\end{figure*}

The detection of blinking lights follows a simple thresholding strategy on the
time differential of images. This approach coupled with decaying
confidence tracking produces satisfactory results for simple motion of robots
and relatively static backgrounds.
Fig. \ref{fig:setupdiagram} shows the
cameras mounted with blinking lights as fiducial markers. The robots shown in
\ref{fig:setupdiagram} are also mounted with ARToolKit\cite{kato1999} fiducial
markers for the comparison experiments.

\section{Experiments}


\label{sec:experiments}

To assess the accuracy of our method we perform a localization experiment in
which we measure how accurately our method can determine the pose of the other
camera. We compare our localization results with the widely used fiducial-based 
pose estimation in ARToolKit \cite{kato1999} and visual egomotion and SfM
framework Bundler \cite{snavely2006photo}.
We also generate a semi-dense reconstruction to compare the mapping
accuracy of our method to that of Bundler. A good quality reconstruction, is a
measure of the accuracy of mutual localization of the two cameras used in the
reconstruction.

\subsection{Localization Experiment}

\paragraph{Setup}
Two turtlebots were set up to face each other. One of the turtlebot was
kept stationary and the other moved in 1 ft increments in an 
X-Z plane (Y-axis is down, Z-axis is along the optical axis of the static camera
and the X-axis is towards the right of the static camera). 
We calculate the rotation error by extracting the rotation angle from the
differential rotation $R_{gt}^\top R_{est}$ as follows: 
\begin{align}
    E_{\theta} = \frac{180}{\pi}\arccos\left(\frac{\Tr(R_{gt}^\top R_{est}) - 1}{2}\right)
    \label{eq:roterr}
\end{align}
where $R_{gt}$ is the ground truth rotation matrix, $R_{est}$ is the
estimated rotation matrix and $\Tr$ is the matrix trace. The translation error
is simply the norm difference between two translation vectors.

\paragraph{Results in comparison with ARToolKit \cite{kato1999}}
The ARToolKit is an open source library for detecting and determining the pose of
fiducial markers from video.
We use a ROS\cite{quigley2009ros} wrapper
-- \href{"http://www.ros.org/wiki/ar_pose"}{\emph{ar\_pose}} --
over ARToolKit for our experiments. We repeat the relative camera localization experiment with the ARToolKit library and compare to our 
results. The results show a \improvementfolds fold improvement in translation error over
Bundler\cite{snavely2006photo}.
%
\begin{table}
  \begin{tabular}{|l|c|c|}
    \hline 
    & Median Trans. error & Median Rotation error\\
    \hline
         ARToolKit\cite{kato1999} & \meanartktranserr & \meanartkroterr \\
    Bundler\cite{snavely2006photo} & \meanbundlertranserr & \meanbundlerroterr \\
 Mutual Localization & \meanmutloctranserr & \meanmutlocroterr \\
    \hline
  \end{tabular}
  \caption{Table showing mean translation and rotation error for ARToolKit,
  Bundler and Mutual Localization}
  \label{tab:artkvsmutloc}
\end{table}

\newcommand{\subfigscale}{0.24\textwidth}
\newcommand{\scaleinsubfig}{1.0\textwidth}
\newcommand{\figwidth}{1.0\columnwidth}
\begin{figure}
  \centering
  \includegraphics[width=\figwidth]{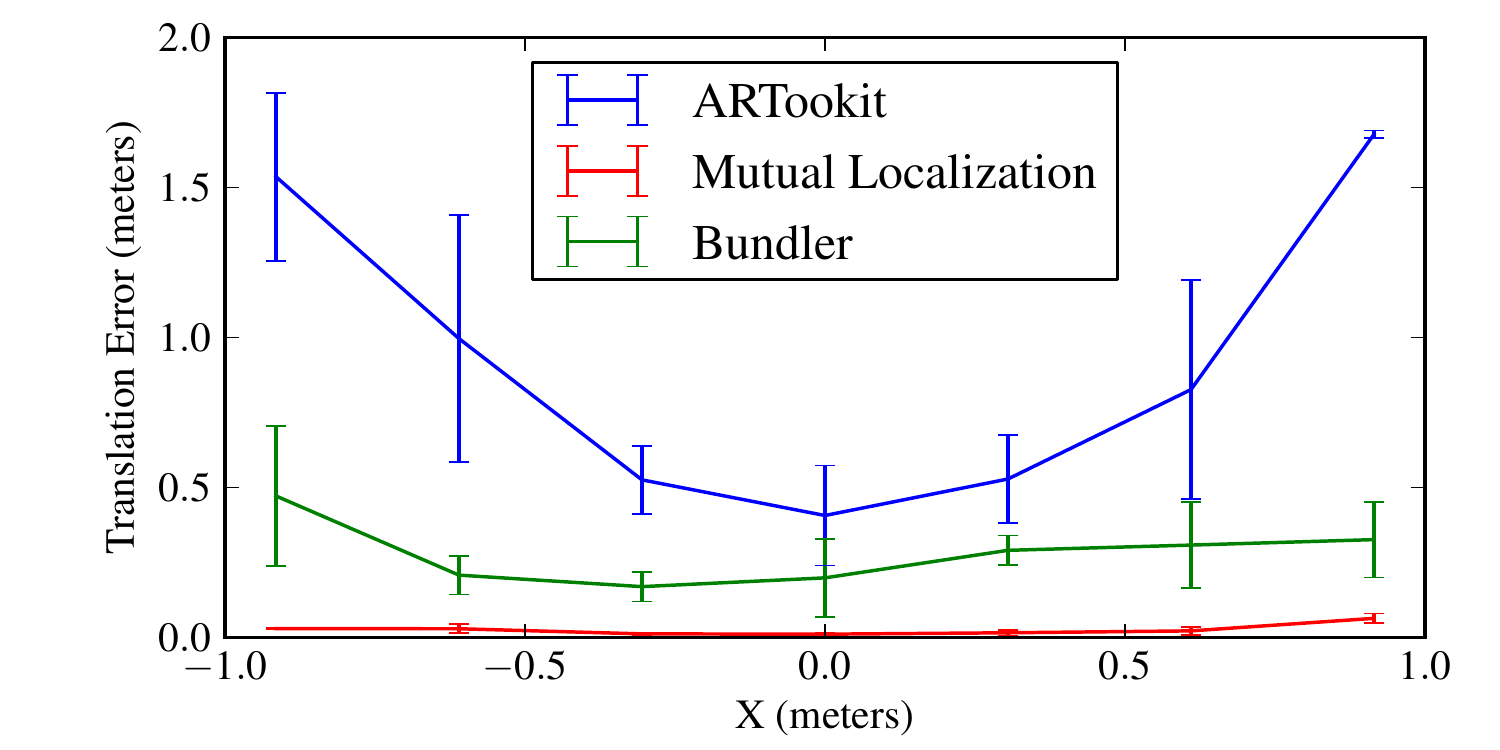}
  \includegraphics[width=\figwidth]{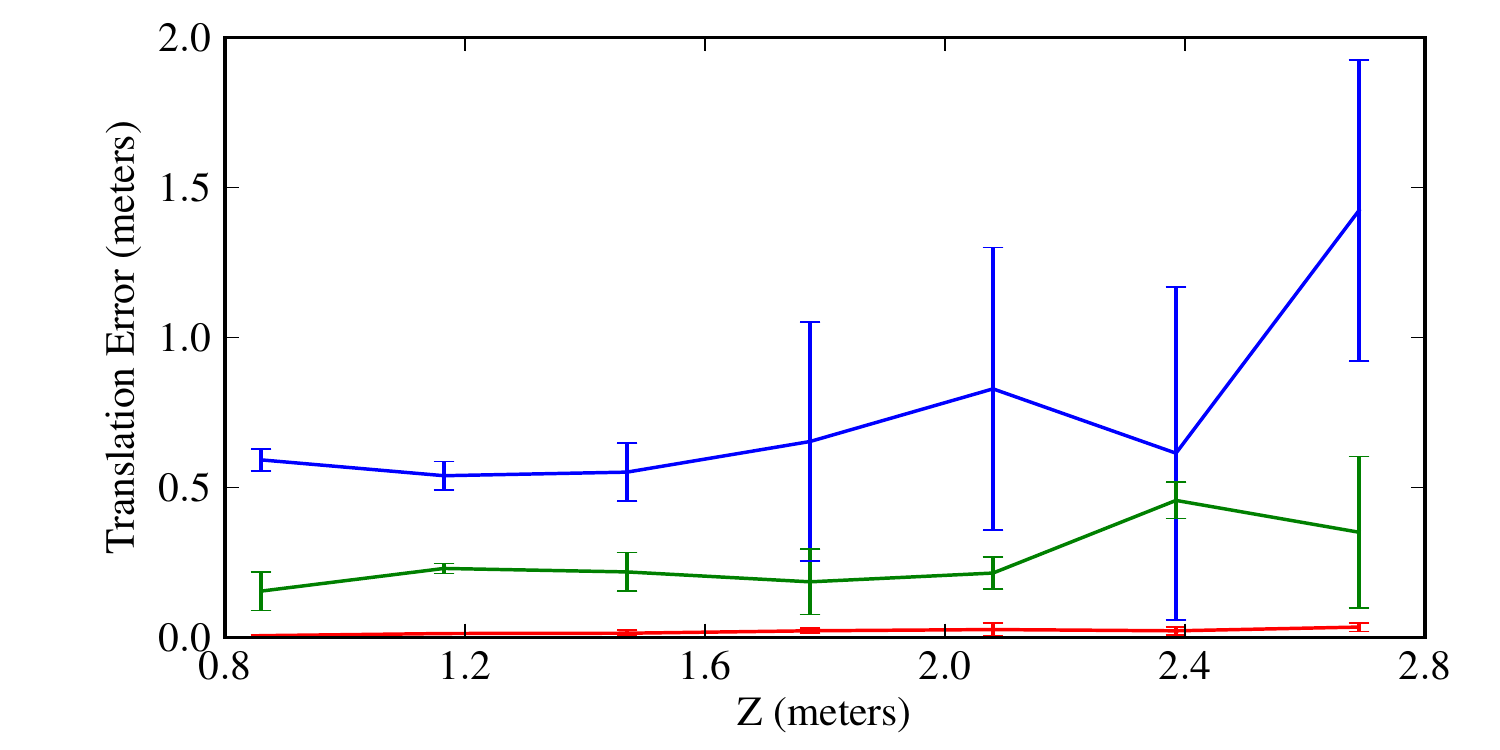}
\caption{Translation error comparison between the ARToolKit and our mutual
localization. The translation error is plotted to ground truth X and Z axis
positions to show how error varies with depth (Z) and lateral (X) movements.
We get better results in localization by a factor of ten. Also note
how the translation error increases with Z-axis (inter-camera separation).}
\label{fig:localizationonly-artkvsmutloc-transerr}
\end{figure}
\begin{figure}
  \centering
  \includegraphics[width=\figwidth]{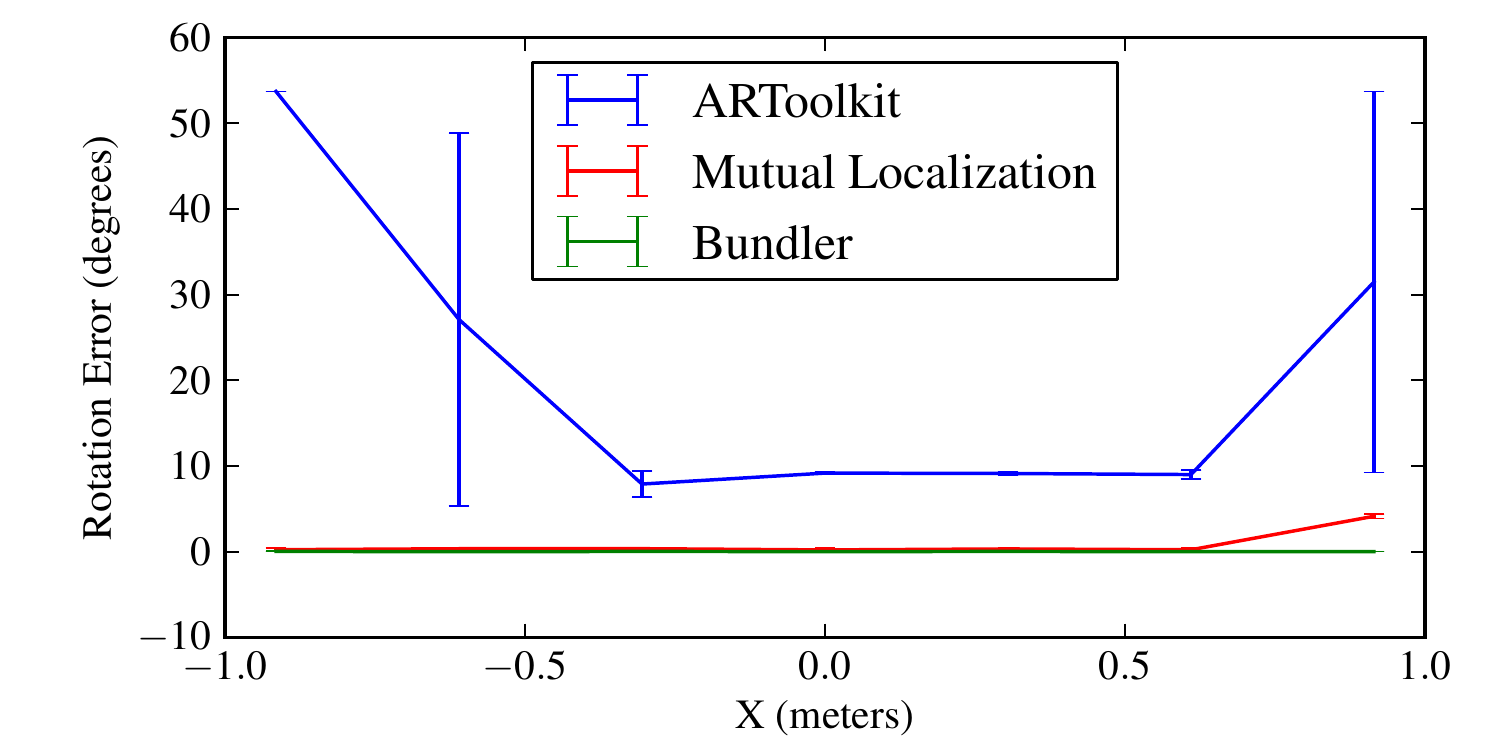}
  \includegraphics[width=\figwidth]{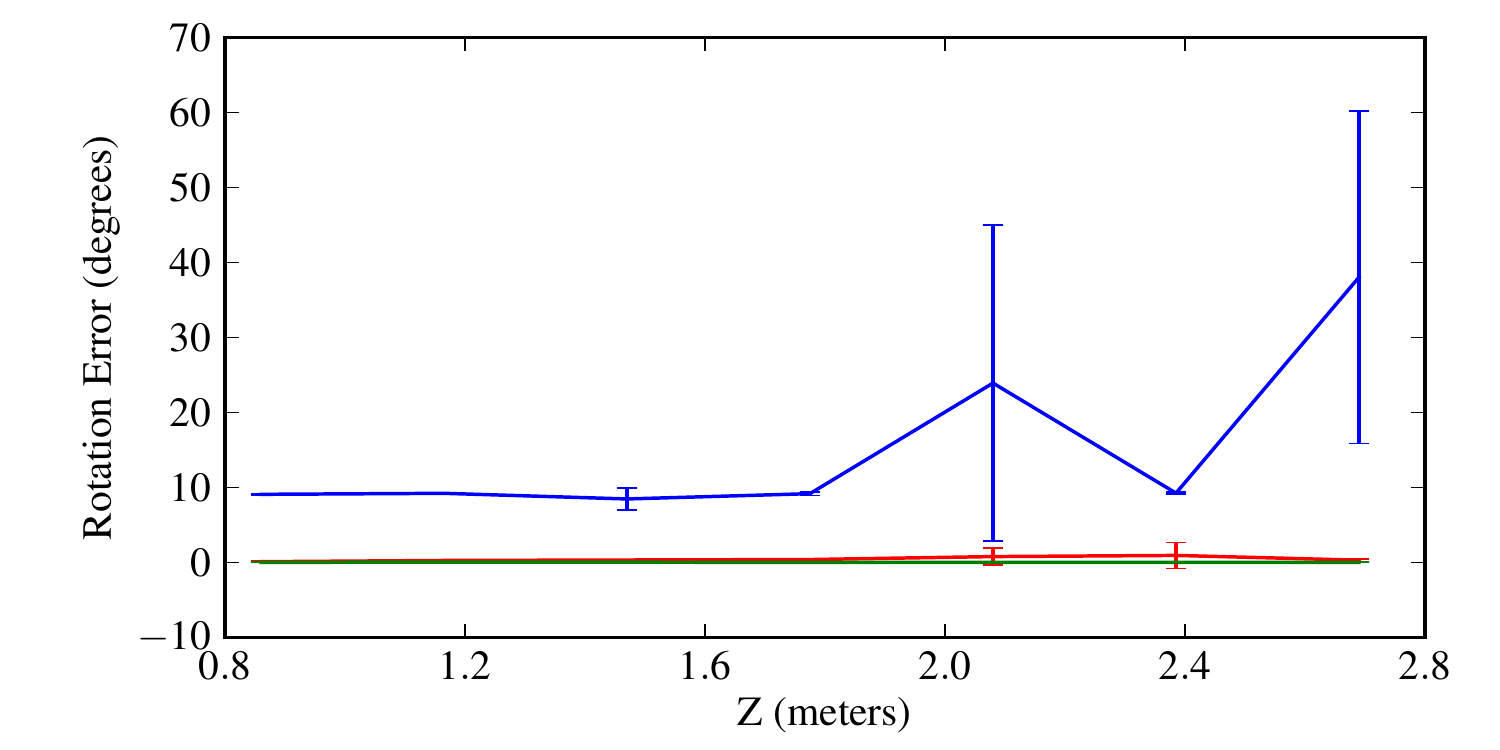}
\caption{Rotation error comparison between the ARToolKit and Mutual
localization. Rotation error decreases with Z-axis (ground truth inter-camera
separation). See \eqref{eq:roterr} for computation of rotation error.}
\label{fig:localizationonly-artkvsmutloc-roterr}
\end{figure}
\subsection{Simulation experiments with noise}
\begin{figure}
  \includegraphics[width=1.0\columnwidth]{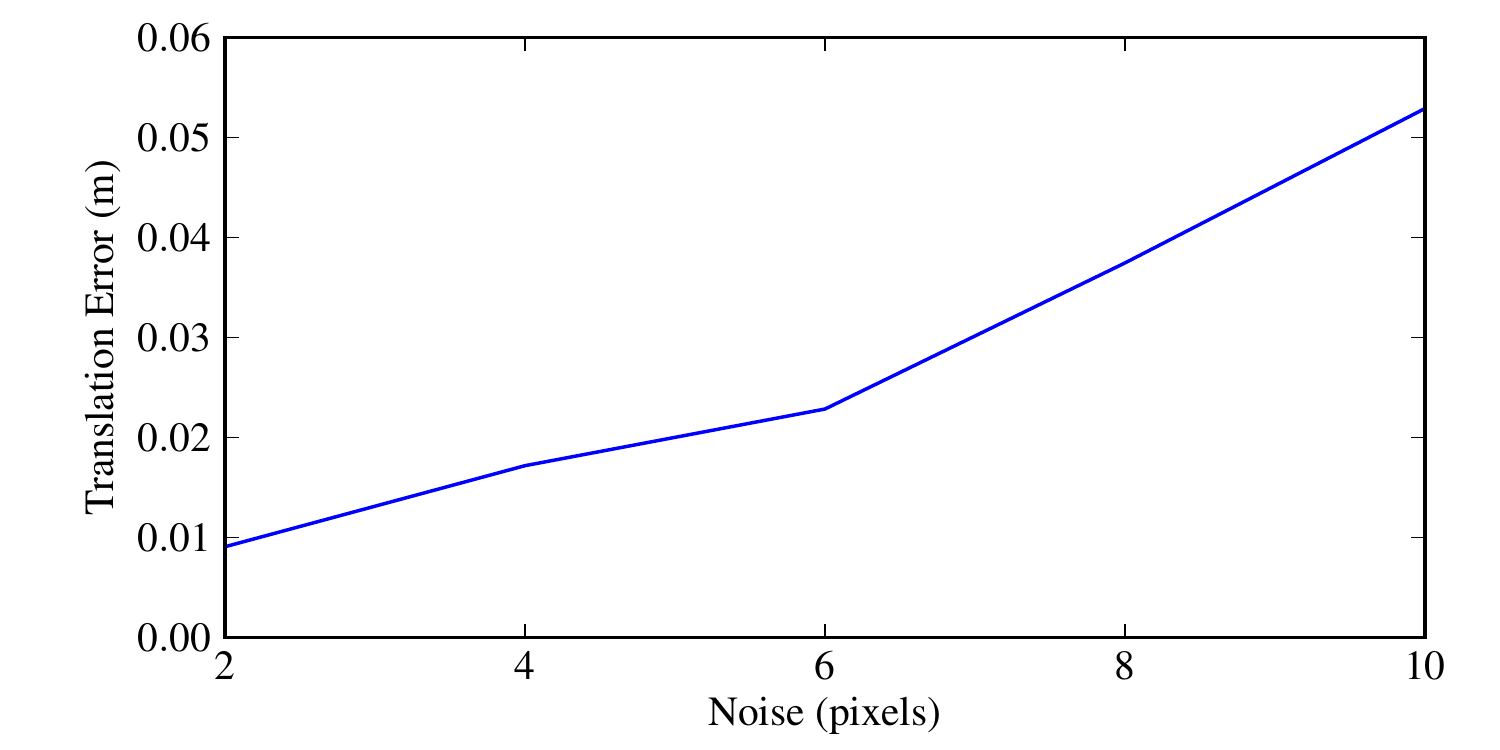}
  \includegraphics[width=1.0\columnwidth]{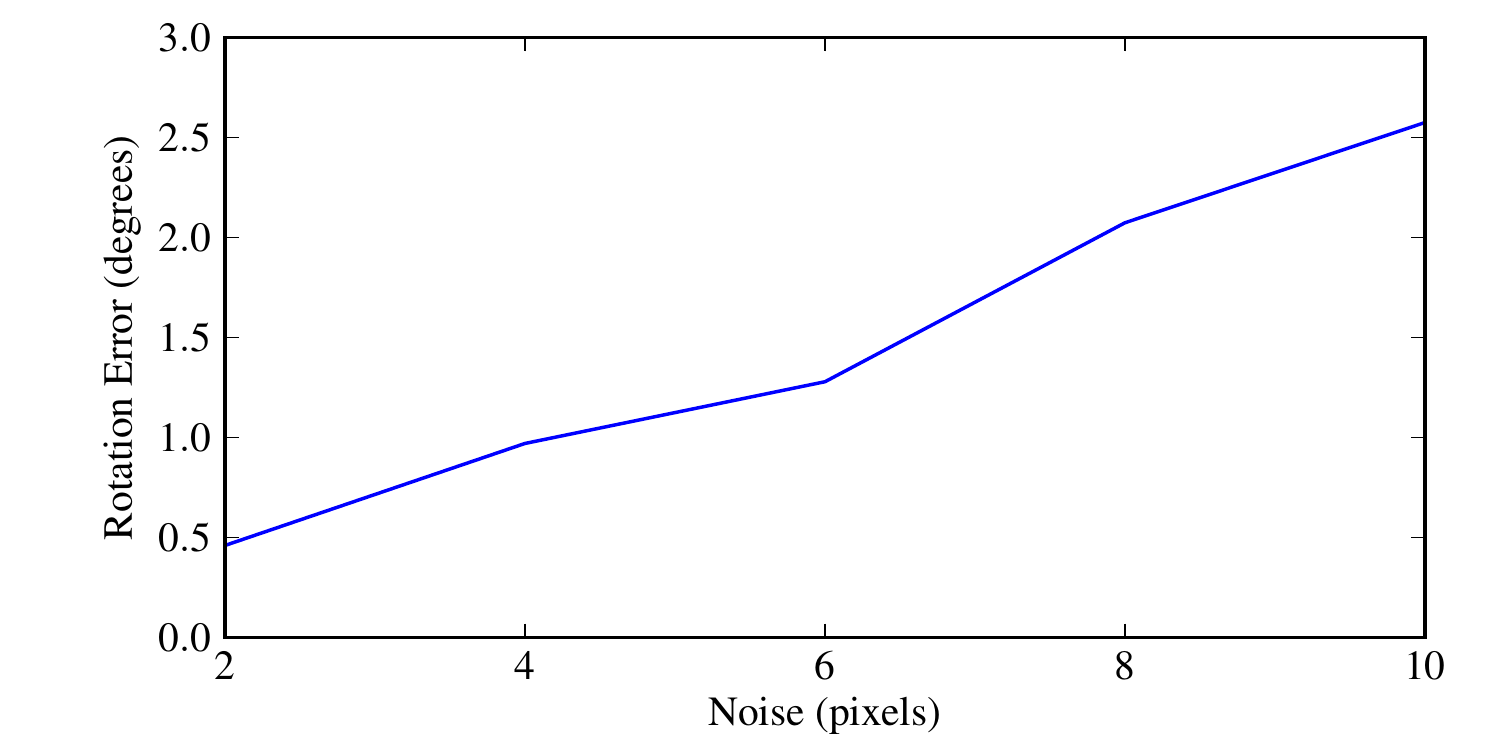}
  \caption{Rotation and translation error as noise is incrementally added to
  the detection of markers.}
  \label{fig:error-vs-noise}
\end{figure}

A simple scene was constructed in Blender to verify the mathematical correctness of
the method. Two cameras were set up in the blender scene along with a target
object 1m from the static camera. Camera images were rendered at a resolution
of 960 $\times$ 540. The markers were simulated as colored balls
that were detected by simple hue based thresholding. The two cameras in the
simulated scene were rotated and translated to cover maximum range of motion.
After detection of the center of the colored balls, zero mean gaussian noise
was added to the detected positions to investigate the noise characteristics
of our method. The experiment was repeated with different values of noise
covariance. Fig \ref{fig:error-vs-noise} shows the translation and rotation
error in the experiment with variation in noise. It can be seen that our
method is robust to noise as it deviates only by 5cm and $2.5^\circ$ when
tested with noise of up to 10 pixels.

\subsection{3D Reconstruction experiment}
\label{sec:3drec}
The position and orientation obtained from our method is inputted into 
the patch based multi-view stereo (PMVS-2) library \cite{furukawa2010accurate} to
obtain a semi-dense reconstruction of an indoor environment. Our reconstruction is 
less noisy when compared to that obtained by Bundler \cite{snavely2006photo}.
Fig. \ref{fig:mappingresults} shows a side-by-side snapshot of the semi-dense map from
Bundler-PMVS and, our method, Mutual Localization-PMVS. To compare the reconstruction
accuracy, we captured the scene as a point cloud with an RGB-D camera (Asus-Xtion).
The Bundler and Mutual Localization output point clouds were
manually aligned (and scaled) to the Asus-Xtion point cloud. We then 
computed the nearest neighbor distance from each point in the Bundler/Mutual
localization point clouds discarding points with nearest neighbors further
than 1m as outliers. With this metric the mean nearest neighbor distance for
our method was 0.176m while that for Bundler was 0.331m.
\renewcommand{\subfigscale}{0.32\textwidth}
\renewcommand{\scaleinsubfig}{1.0\textwidth}
\begin{figure*}
  \begin{subfigure}[b]{\subfigscale}
    \includegraphics[width=\scaleinsubfig]{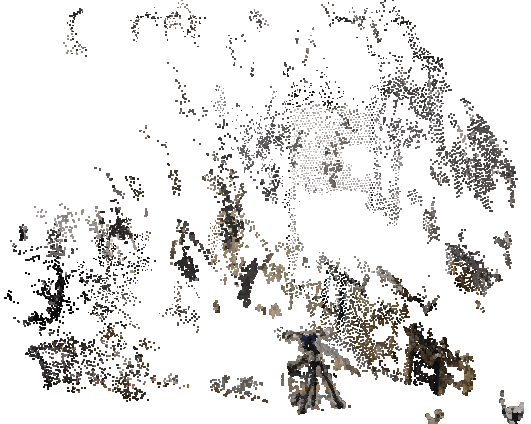}
    \caption{Bundler-PMVS}
  \end{subfigure}
  \begin{subfigure}[b]{\subfigscale}
    \includegraphics[width=\scaleinsubfig]{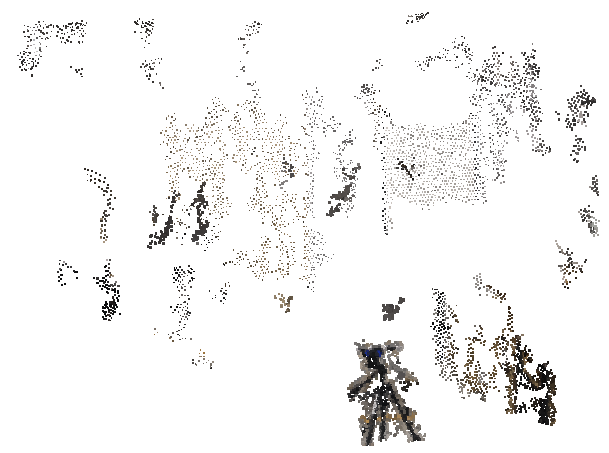}
    \caption{Mutual Localization-PMVS}
  \end{subfigure}
  \begin{subfigure}[b]{\subfigscale}
    \includegraphics[width=\scaleinsubfig]{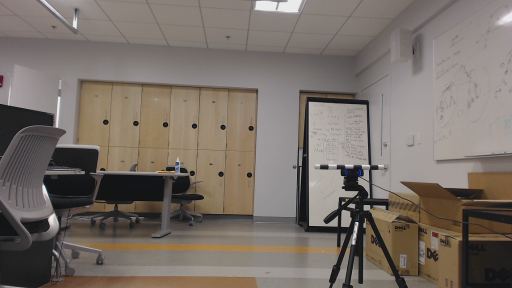}
  \caption{Actual scene}
  \end{subfigure}
  \caption{The semi-dense reconstruction produced by our method, Mutual 
  Localization, is less noisy (0.18m) when compared to that produced by
Bundler (0.33m).}
  \label{fig:mappingresults}
\end{figure*}

%



\section{Conclusion}

We have developed a method to cooperatively localize two cameras using fiducial
markers on the cameras in sensor-specific coordinate frames, obviating the
common assumption of sensor egomotion. We have compared our results with the
ARToolKit showing that our method can localize significantly more accurately,
with a \improvementfolds fold error reduction observed in our experiments.
We have also demonstrated how the cooperative localization can be used as an
input for 3D reconstruction of unknown environments, and find better
accuracy (0.18m versus 0.33m) than the visual egomotion-based Bundler method.
We plan to build on this work and apply it to multiple robots for cooperative mapping.
Though we achieve reasonable accuracy, we believe we can improve the accuracy of our method 
by improving camera calibration and measurement of the fiducial marker locations with
respect to the camera optical center.
We will release the source code (open-source) for our method upon publication.

\section*{Acknowledgments}
This material is based upon work partially supported by the Federal Highway
Administration under Cooperative Agreement No. DTFH61-07-H-00023, the Army
Research Office (W911NF-11-1-0090) and the National Science Foundation CAREER
grant (IIS-0845282). Any opinions, findings, conclusions or recommendations are
those of the authors and do not necessarily reflect the views of the FHWA, ARO,
or NSF.

\bibliographystyle{IEEEtran}
\bibliography{paper}

\end{document}